\pdfoutput=1
\documentclass[11pt]{article}

\usepackage{acl}

\usepackage{times}
\usepackage{latexsym}
\usepackage{multicol}
\usepackage{booktabs}
\usepackage[fleqn]{amsmath}
\usepackage{multirow}
\usepackage[normalem]{ulem}
\usepackage{hyperref}
\usepackage{graphicx}
\usepackage{subfigure}
\usepackage{fixltx2e}
\usepackage{algorithm}
\usepackage{algorithmic}
\usepackage{setspace}
\usepackage{soul}
\usepackage{amsmath}
\usepackage{amsthm}
\usepackage{colortbl}
\usepackage{xcolor}
\usepackage{pifont}
\newcommand{\cmark}{\ding{51}}%
\newcommand{\xmark}{\ding{55}}%

\usepackage[T1]{fontenc}
\usepackage[utf8]{inputenc}

\usepackage{microtype}

\usepackage{inconsolata}

\usepackage{graphicx}

\title{UniConv: Unifying Retrieval and Response Generation for Large Language Models in Conversations}

\author{Fengran Mo$^1$\thanks{Work done when interned in Amazon.}, Yifan Gao$^2$, Chuan Meng$^{3*}$, Xin Liu$^2$, Zhuofeng Wu$^2$, Kelong Mao$^{4*}$\\ \textbf{Zhengyang Wang}$^2$, \textbf{Pei Chen}$^2$, \textbf{Zheng Li}$^2$, \textbf{Xian Li}$^2$, \textbf{Bing Yin}$^2$, \textbf{Meng Jiang}$^5$\\
$^1$University of Montreal; 
$^2$Amazon.com;
$^3$University of Amsterdam\\ 
$^4$Renmin University;
$^5$University of Notre Dame\\ 
\texttt{fengran.mo@umontreal.ca, yifangao@amazon.com, mjiang2@nd.edu} \\
}

\begin{document}
\maketitle
\begin{abstract}
The rapid advancement of conversational search systems revolutionizes how information is accessed by enabling the multi-turn interaction between the user and the system. 
Existing conversational search systems are usually built with two different models. 
This separation restricts the system from leveraging the intrinsic knowledge of the models simultaneously, which cannot ensure the effectiveness of retrieval benefiting the generation. 
The existing studies for developing unified models cannot fully address the aspects of understanding conversational context, managing retrieval independently, and generating responses.
In this paper, we explore how to unify dense retrieval and response generation for large language models in conversation. 
We conduct joint fine-tuning with different objectives and design two mechanisms to reduce the inconsistency risks while mitigating data discrepancy. The evaluations on five conversational search datasets demonstrate that our unified model can mutually improve both tasks and outperform the existing baselines.
\end{abstract}

\section{Introduction}
% The meaning and development of Convsational Search
The rapid advancement of conversational search systems revolutionizes how information is accessed by enabling the multi-turn interaction between the user and the system~\cite{zamani2023conversational}.
With the recent advances of large language models (LLMs), commercial conversational AI search engines, such as Perplexity.ai and SearchGPT\footnote{Perplexity.ai: \url{https://www.perplexity.ai/}, SearchGPT: \url{https://searchgpt.com/}}, have been deployed with increasing attraction. 

%users and significant growth.

% Two separated component: retrieval and generation in conversational search. Indicate the disadvantage of this framework and why we need unified model
Existing conversational search systems are usually composed of two different models: a retriever and a generator~\cite{gao2022neural,mo2024survey}, which are trained and inferred separately. The retriever aims to identify the relevant passages by understanding conversational queries, while the generator crafts the final response for the information-seeking goal. The deployment of separate models in the whole pipeline induces the problems in two folds: \textit{i)} The separation restricts the system from leveraging the model's intrinsic knowledge simultaneously, which raises the risk of lacking correlation with the performance of both tasks, leading to inconsistent results, i.e., the effectiveness of retrieval might not always benefit response generation~\cite{salemi2024evaluating}; and \textit{ii)} Deploying and maintaining two distinct models adds extra hardware requirements and increases maintenance expenses~\cite{zhang2024onegen}. 
An intuitive solution is to develop a unified model that acts as a retriever and generator in conversational scenarios. This model is expected to mutually improve retrieval and generation performance through seamless integration with end-to-end optimization.

% The existing studies for unifying, and illustrate their flaws. Motivate our idea by introducing the challenging of develop unified model for conversational search.
Recent existing studies have demonstrated the feasibility of developing LLM-based unified models in conversational question answering, involving open-domain retrieval\footnote{In this paper, the retrieval denotes retrieving information from a large external collection as an open-domain setting, rather than only identifying specific pieces from the initial search results with limited top-k candidates similar to ranking.} and response generation. 
However, these systems can only address two aspects of understanding conversational context, managing retrieval independently, or generating responses, as illustrated in Table~\ref{table:comparison}.
Among them, RepLLaMA~\cite{ma2024fine} and E5~\cite{wang2024improving} successfully implement generative LLMs for retrieval tasks and ChatRetriever~\cite{mao2024chatretriever} further adapt it to conversational scenarios. However, the fine-tuning for retrieval objectives leads to the collapse of generation ability in these systems.
The RankRAG~\cite{yu2024rankrag} and ChatQA~\cite{liu2024chatqa} enable the system to exploit a more accurate input context for the generator to produce better responses to user queries.
However, they should rely on an additional retriever to address retrieval needs.
To develop a unified model capable of handling both retrieval and generation tasks, GRIT~\cite{muennighoff2024generative} attempts to train an LLM with distinguished instructions but it is not designed for conversational applications.

\begin{table}[t]
    \centering
    \scalebox{0.85}{
    \begin{tabular}{lccc}
    \toprule
    \textbf{System} &
    \small\textbf{Conv.} & \small\textbf{Ret.} & \small\textbf{Gen.} \\ 
    \midrule
    RepLLaMA~\cite{ma2024fine} & \xmark & \cmark & \xmark \\
    E5~\cite{wang2024improving} & \xmark & \cmark & \xmark \\
    ChatRetriever~\cite{mao2024chatretriever} & \cmark & \cmark & \xmark \\
    RankRAG~\cite{yu2024rankrag} & \cmark & \xmark & \cmark \\
    ChatQA~\cite{liu2024chatqa} & \cmark & \xmark & \cmark \\
    %OneGen~\cite{zhang2024onegen} & \xmark & \xmark & \cmark \\ in parallel work 
    GRIT~\cite{muennighoff2024generative} & \xmark & \cmark & \cmark \\
    \midrule
    Our UniConv & \cmark & \cmark & \cmark\\
    \bottomrule
    \end{tabular}}
    \caption{The functionality comparison between ours and existing systems, including whether support to conversational scenario (Conv.), first-stage retrieval (Ret.), and response generation (Gen.).}
    \label{table:comparison}
%\vspace{-2ex}
\end{table}

% Introduce our methodology and observation.
To address the limitations of previous studies, we propose \textbf{UniConv}, a unified LLM-based model to investigate the feasibility of handling both retrieval and response generation in conversation. 
To achieve this, we inherit the training data selected by ChatRetriever~\cite{mao2024chatretriever} to adapt LLM to serve as a powerful conversational dense retriever. To improve the response generation ability while fine-tuning dense retrieval, we design a context identification instruction mechanism as part of the learning objective. This mechanism is designed to seamlessly integrate retrieved information into the response generation process.
Additionally, we identify a discrepancy in previous training data: the same data format is applied to different learning objectives, which does not align well with the distinct output requirements of retrieval and generation tasks.
To mitigate this issue, we include additional well-formatted conversational search data for model fine-tuning.
We conduct extensive evaluations on five widely used datasets, where UniConv demonstrates strong generalization capabilities for representing complex conversational sessions in dense retrieval, along with robust generation abilities for crafting responses.
Moreover, UniConv achieves better seamless consistency between retrieval and its augmentation for response generation in terms of effectiveness and reliability compared to non-unified models.

Our contributions can be summarized as follows:

(1) We investigate the feasibility of developing a unified LLM for conversational search and propose our UniConv model for better unification.

(2) We design two mechanisms to improve the seamless consistency between conversational retrieval and its augmented response generation while addressing the issue of data discrepancy.

(3) We conduct extensive experiments to evaluate UniConv on information-seeking conversations across various settings, comparing it against several strong baselines. Its superior performance in both conversational dense retrieval and response generation highlights its remarkable effectiveness.

\section{Related Work}
\noindent \textbf{Conversational Retrieval.}
Conversational retrieval aims to identify the relevant passages to satisfy users' information needs through multi-turn interaction~\citep{meng2025bridging,mo2025conversational}. The main challenge lies in enabling the system to understand the real user search intents expressed in context-dependent queries. 
The literature outlines two main approaches to achieve the retrieval goal: \textit{i)} conversational query rewriting~\cite{voskarides2020query,wu2022conqrr,mo2023convgqr,mo2023learning,mao2023large,mao2023search,ye2023enhancing,jang2023itercqr,mo2024chiq,mo2024leverage,lai2024adacqr} that decomposes the conversational retrieval into a rewrite-then-retrieval pipeline and \textit{ii)} conversational dense retrieval~\cite{qu2020open,yu2021few,lin2021contextualized,kim2022saving,mao2022convtrans,mao2023learning,jin2023instructor,mo2024aligning,mo2024history,mo2024convsdg,lupart2024disco} that directly encode the whole conversational session to perform end-to-end dense retrieval. \\

\noindent \textbf{Conversational Response Generation.} 
Conversational response generation aims to synthesize information from the top-retrieved passages into a single response~\cite{meng2020refnet,meng2020dukenet,meng2021initiative,ren2021conversations,cheng2024coral,li2024mosaic}. Different from single-turn retrieval-augmented generation (RAG)~\cite{lewis2020retrieval,asai2023self,mao2024rag,zhang2025entropy}, which only needs to consider the given stand-alone query with its associated retrieved results for response generation, a conversational response generator~\cite{ye2024boosting} requires modeling conversational turn dependency and understanding the context-depend query and search results. \\

\noindent \textbf{LLM-based Retrieval.} To explore the potential of LLMs in retrieval tasks, some existing studies~\cite{wang2024improving,ma2024fine} attempt to follow the observed scaling law~\cite{kaplan2020scaling} in search model~\cite{ni2022large} by replacing the backbone model from the small ones (e.g., BERT-base~\cite{devlin2019bert} and T5-base~\cite{raffel2020exploring}) into the generative LLMs (e.g., Mistral~\cite{arxiv23_mistrial} and LLaMa~\cite{arxiv23_llama2}). They keep the training paradigm similar to DPR~\cite{karpukhin2020dense} using relevance judgments as supervision signals while changing the representation of queries and passages different from the ones in encoder-based models. \\

\noindent \textbf{Unified LLMs for Retrieval and Generation.} 
The motivation to develop unified LLMs for retrieval and generation is to attempt to mutually leverage the intrinsic knowledge from both sides to improve the model's general multi-task ability and reduce cost.
To this end, ~\citet{muennighoff2024generative} propose GRIT, to train LLMs to handle both generative and retrieval tasks by distinguishing between them through instructions, and ~\citet{li2024unigen} design a unified framework based on generative retrieval and open-domain question answering.
Then,~\citet{yu2024rankrag} propose RankRAG, which unifies the re-ranking and generation through simultaneously instructing the LLMs on context ranking and answer generation. 
However, they cannot address multi-turn scenarios due to a lack of conversational adaptation. 
In a conversational setting, \citet{mao2024chatretriever} and \citet{liu2024chatqa} fine-tune LLMs specifically for conversational retrieval and response generation tasks, respectively, but these adaptations do not preserve the model's ability to perform both functions concurrently. 
Recently, a parallel study, OneGen~\cite{zhang2024onegen}, propose unifying the traditionally separate training approaches for generation and retrieval by incorporating retrieval tokens generated in an autoregressive manner. However, it cannot follow conversational context and independently handle retrieval tasks. \\

%\noindent\textbf{Our Goal.} 
%We aim to develop a unified LLM-based model capable of handling both retrieval and generation in conversational scenarios, which is not well-explored in aforementioned existing literature.

\noindent\textbf{Our Goal} \quad is to develop a unified LLM-based model capable of handling both retrieval and generation in conversation, a scenario that has not been extensively explored in the existing literature.

\section{Methodology}

\subsection{Task Formulation}
The task is to establish a unified model which can handle both conversational retrieval and response generation. Formally, given a conversational session that contains $n-1$ historical turns $\mathcal{H}_n=\{q_i,r_i\}_{i=1}^{n-1}$ and current query $q_n$, the unified model $\mathcal{M}$ is expected to act as a retriever to identify the relevant passages $\mathcal{P}_n$ from a large collection $\mathcal{C}$ and also act as a generator to produce a response $r_n$ on top of external knowledge $\mathcal{P}_n$ to satisfy the information needs in $q_n$. 
Thus, the unified model $\mathcal{M}$ is required to handle the multi-turn session query to retrieve the relevant passages as $\mathcal{P}_n=\mathcal{M}(q_n,\mathcal{H}_n)$, and generate the response as $r_n=\mathcal{M}(q_n,\mathcal{H}_n,\mathcal{P}_n)$. In our setting, the unified model $\mathcal{M}$ is a generative LLM with decoder-only architecture.

\begin{figure}[t]
\centering
%\vspace{-2ex}
\includegraphics[width=1\linewidth]{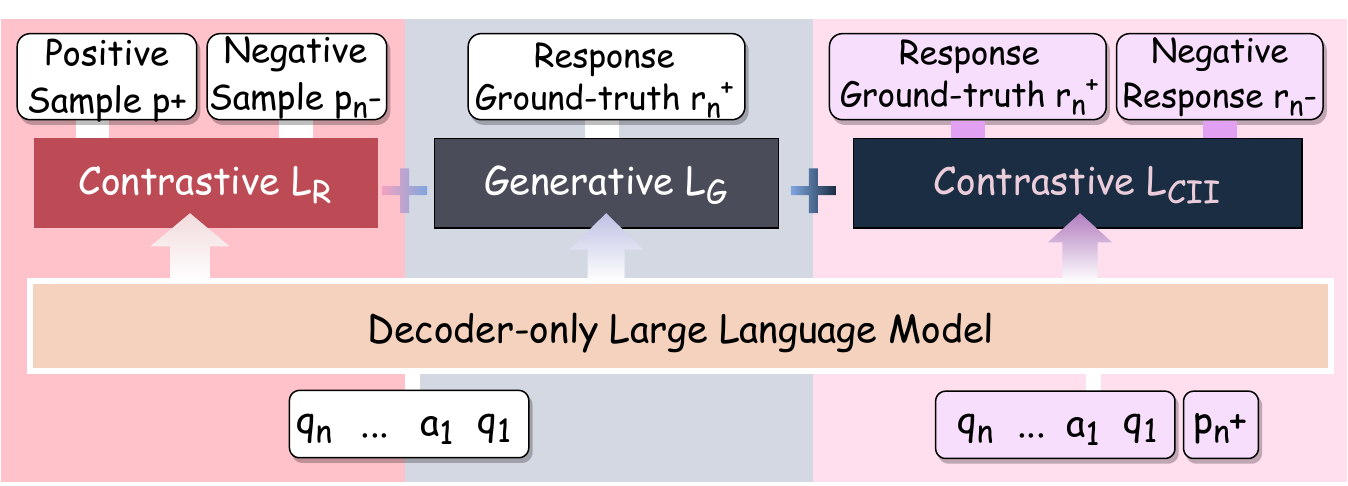}
\caption{Overview of our UniConv. Three learning objectives are designed with various input and supervision signals for retrieval and generation in conversation.}
\label{fig: overview}
%\vspace{-2ex}
\end{figure}

\subsection{Generative Language Models for Conversational Search}
The overview of our proposed UniConv is illustrated in Figure~\ref{fig: overview}, which consists of three parts, including various learning objectives toward conversational retrieval (Sec.~\ref{subsec: CDR}), conversational response generation (Sec.~\ref{subsec: CRG}), and context identification instruction (Sec.~\ref{subsec: CII}). We describe each component as follows.

\subsubsection{Conversational Dense Retrieval}
\label{subsec: CDR}
The common practice for dense retrieval fine-tuning is the paradigm of DPR~\cite{karpukhin2020dense}, which leverages a bi-directional encoder-only model to encode the queries and passages separately on top of a bi-encoder architecture. Then, the first sequence token \texttt{[CLS]} is employed as the text representation for similarity calculation. 
When the backbone model $\mathcal{M}$ turns into generative ones with uni-directional decoder-only architecture, e.g., LLaMA, the adaption is to form the representation $\mathcal{V}_{\text{seq}}$ using an appended end-of-sequence token \texttt{</s>} to both the queries and passages~\cite{ma2024fine} as 
$\mathcal{V}_{\text{seq}} = \mathcal{M}(\cdot)[-1]$.

To adapt the conversational scenario, the input query for each turn $n$ is reformulated as $q_n^{\prime}=q_n \circ \mathcal{H}_n$ by concatenating the context of the previous turn. Then it is vectorized with candidate passages $p_n$ by the model $\mathcal{M}$ and calculate their similarity $\mathcal{S}\left(q^{\prime}_{n}, p_{n}\right)=<\mathcal{V}_{q^{\prime}_{n}}, \mathcal{V}_{p}>$ via dot product.
With the established representation, contrastive learning with InfoNCE loss is used for end-to-end conversational dense retrievers optimization as
%The final training objective for conversational dense retrievers is 
$$
\mathcal{L}_{\text{R}} = - \log  \frac{e^{\mathcal{S}\left(q^{\prime}_{n}, p^{+}_{n}\right)}} {e^{\mathcal{S}\left(q^{\prime}_{n}, p^{+}_{n}\right)} + \sum_{\mathcal{P}^{-}_n \in \{\mathcal{P}_{N}\}} e^{\mathcal{S}\left(q^{\prime}_{n}, \mathcal{P}^{-}_n\right)}}
$$

\subsubsection{Conversational Response Generation}
\label{subsec: CRG}
For information-seeking response generation in the conversation, the generator shares the same comprised query input $q_n^{\prime}$ as the retriever and is required to maintain the generation ability while learning for retrieval. To enhance the robustness of the generator, we inherit the idea of Seq2Seq~\cite{sutskever2014sequence}, enabling the model to only be conditional on the representation of the input query $\mathcal{V}_{q^{\prime}_{n}}$ rather than attention on all previous input and generated tokens. This is achieved by applying the session-masked technique in~\cite{mao2024chatretriever} and the training objective to generate the ground-truth for turn $n$ is shown below, where $|r_n| = T$.
$$
\mathcal{L}_{\text{G}} = -\frac{1}{T}\sum_{i=1}^{T}\log p(r_n^{i}|r_n^{1}, ..., r_{n}^{i-1}, \mathcal{V}_{q^{\prime}_{n}}) 
$$

\subsection{Context Identification Instruction}
\label{subsec: CII}
During the inference phase with the retrieval-augmented setting, the model input is usually the query together with the retrieved evidence serving as the main part of the instruction, where the model is expected to generate the response grounding on the useful piece of the retrieved evidence. 
Since the relevant passage and ground-truth response used as supervision signals are separately conducted during the training phase within the unified model, potential inconsistency risk might occur~\cite{yu2024rankrag}.
To this end, we design a context identification instruction to help the model implicitly identify the useful passage during the fine-tuning, which is consistent with the input instruction format of inference. This is achieved by combining the query with the positive passage in the same sequence and using different responses as contrastive samples as
$$
\mathcal{L}_{\text{CII}} = - \log  \frac{e^{\mathcal{S}\left(q^{\prime}_{n} \circ p^{+}_{n}, r_n^+\right)}} {e^{\mathcal{S}\left(q^{\prime}_{n} \circ p^{+}_{n}, r_n^+\right)} + \sum\limits_{{r}^{-}_n \in \{{r}\}} e^{\mathcal{S}\left(q^{\prime}_{n} \circ p^{+}_{n}, r_n^-\right)}}
$$

\subsection{Data Discrepancy Mitigation}
% include conversational search data
To equip the LLMs with conversational dense retrieval capability,~\citet{mao2024chatretriever} leverage the ad-hoc search data with relevant query-passage pairs and instructional conversation with multi-turn query-response pairs to enable the model to obtain retrieval and conversational context understanding ability.
In practice, their implementation utilizes each turn’s response $r_n^+$ in the conversation dataset as the relevant passage $p_n^{+}$ and each ad-hoc query's corresponding relevant passage $p_n^{+}$ as the ground-truth $r_n^+$ for retrieval and generation fine-tuning, respectively. 
However, a unified model should have different outputs for conversational retrieval (e.g., rank-list) and generation (e.g., synthesized response), whose requirement is not exactly matched with the fine-tuned data form in existing studies and thus might lead to sub-optimal results.
A more practical way is to ensure each data sample includes both the relevant passage $p_n^+$ and the corresponding ground-truth response $r_n^+$ as supervision signals for the given query turn $q_n$. Then, the model can learn the consistency from the various targets between retrieval and generation. Thus, we include the conversational search data~\cite{adlakha2022topiocqa} to meet this requirement to mitigate the data discrepancy issue. Another alternative is to construct synthetic data~\cite{liu2024chatqa} with well-formed signals, which is not the focus of our paper. %and we left it for future study.

\subsection{Training and Inference}
For the training phase, we integrate the conversational dense retrieval, retrieval-augmented response generation, and the context identification instruction tuning to form the training objective $\mathcal{L}$ of our unified model as Eq.~\ref{eq: loss}, 
%with ad-hoc search, instructional conversation, and conversational search datasets for data discrepancy mitigation, 
where $\alpha$ is a hyper-parameter to control the fine-tuning effect. For the inference phase, we use the same fine-tuned model to perform retrieval to produce a top-$k$ rank list and generation to produce a response within zero-shot and RAG settings under conversational scenarios.
\begin{equation}
    \mathcal{L} = \mathcal{L}_{\text{R}} + \mathcal{L}_{\text{G}} + \alpha \mathcal{L}_{\text{CII}}
\label{eq: loss}
\end{equation}

\section{Experiments}

\subsection{Experimental Setup}
\noindent \textbf{Evaluation Datasets and Metric.} 
We conduct the main evaluation on four widely-used conversational search datasets, including TopiOCQA~\cite{adlakha2022topiocqa}, QReCC~\cite{anantha2021open}, OR-QuAC~\cite{qu2020open}, and INSCIT~\cite{wu2023inscit}. 
Each of them contains the gold standard for passage retrieval and response generation. 
Besides, FaithDial~\cite{dziri2022faithdial}, an information-seeking dialogue benchmark, and TopiOCQA are used for evaluating the reliability of the generated content via the given evidence/rationale.
The statistics and details of the datasets are provided in Appendix~\ref{sec: appendix_dataset}.
We use NDCG@3, Recall@10, and F1 to evaluate the retrieval and generation performance to conduct a fair comparison with baselines. \\
%following the existing studies~\cite{mao2024chatretriever,liu2024chatqa}. 

\noindent \textbf{Training data.}
We use the ad-hoc search dataset MSMARCO~\cite{bajaj2016ms}, the \textit{The Question About the World} subset of the instructional conversation dataset UltraChat~\cite{ding2023enhancing}, and the whole conversational search dataset TopiOCQA for fine-tuning the unified model. \\

\noindent \textbf{Baselines.}
We compare our methods with various conversational retrieval and response generation baselines. 
For the retrieval phase, we compete with the most effective conversational dense retrieval (CDR) systems based on small language models (SLMs), including ConvDR~\cite{yu2021few}, Conv-ANCE~\cite{mao2023learning}, and QRACDR~\cite{mo2024aligning} and most recently LLM-based approaches, including RepLLaMA~\cite{ma2024fine}, E5~\cite{wang2024improving}, (Conv-)GRIT~\cite{muennighoff2024generative}, and ChatRetriever~\cite{mao2024chatretriever}. 
The GRIT is the only system that can handle both retrieval and generation tasks, and its variant Conv-GRIT is fine-tuned for conversation on the same setting as ours. Besides, the compared baselines also contain the methods based on conversational query rewriting (CQR) on top of LLMs, including the ones without fine-tuning (LLM-Aided~\cite{ye2023enhancing}, LLM4CS~\cite{mao2023large}, and CHIQ~\cite{mo2024chiq}) and with fine-tuning (RETPO~\cite{yoon2024ask}).

In the response generation phase, we conduct the comparison under zero-shot and RAG settings. 
For the zero-shot manner, we include GRIT and its variants Conv-GRIT and three powerful LLMs: Mistral, Claude~\cite{claude}, and ChatGPT~\cite{chatgpt3_5}. 
For the RAG setting, to make the results comparable, we employ Mistral-2-7B-chat as the generator with two typical conversational dense retrievers Conv-ANCE and Chatretriever, and keep the Conv-GRIT on the same workflow as our UniConv, i.e., using the same model for both tasks. More details about the baseline methods are described in Appendix~\ref{sec: appendix_baseline}. \\

\noindent \textbf{Implementation Details.}
We initialize UniConv with Mistral-2-7B-chat, which can
be also applied on top of any generative models. We train it on eight 40G A100 GPUs using LoRA~\cite{hu2022lora} with a maximum input sequence length of 1024 for query and 384 for passages and responses.
The training process involves one epoch with a learning rate of 1e-4, a gradient accumulation of 4 steps, a batch size of 32, and in-batch negatives per sample. The $\alpha$ for loss balance is set to 0.5.
During the inference stage, we deploy Faiss~\citep{johnson2019billion} for the dense retrieval, set the maximum length as 128, and use top-10 retrieved passages for the response generation.
For baseline comparisons, we adhere to the implementation settings specified in their original papers. The LLM-based CQR and the SLM-based CDR methods are based on ANCE dense retriever~\cite{xiong2020approximate}. 

\begin{figure}[h]
\centering
%\vspace{-2ex}
\includegraphics[width=1\linewidth]{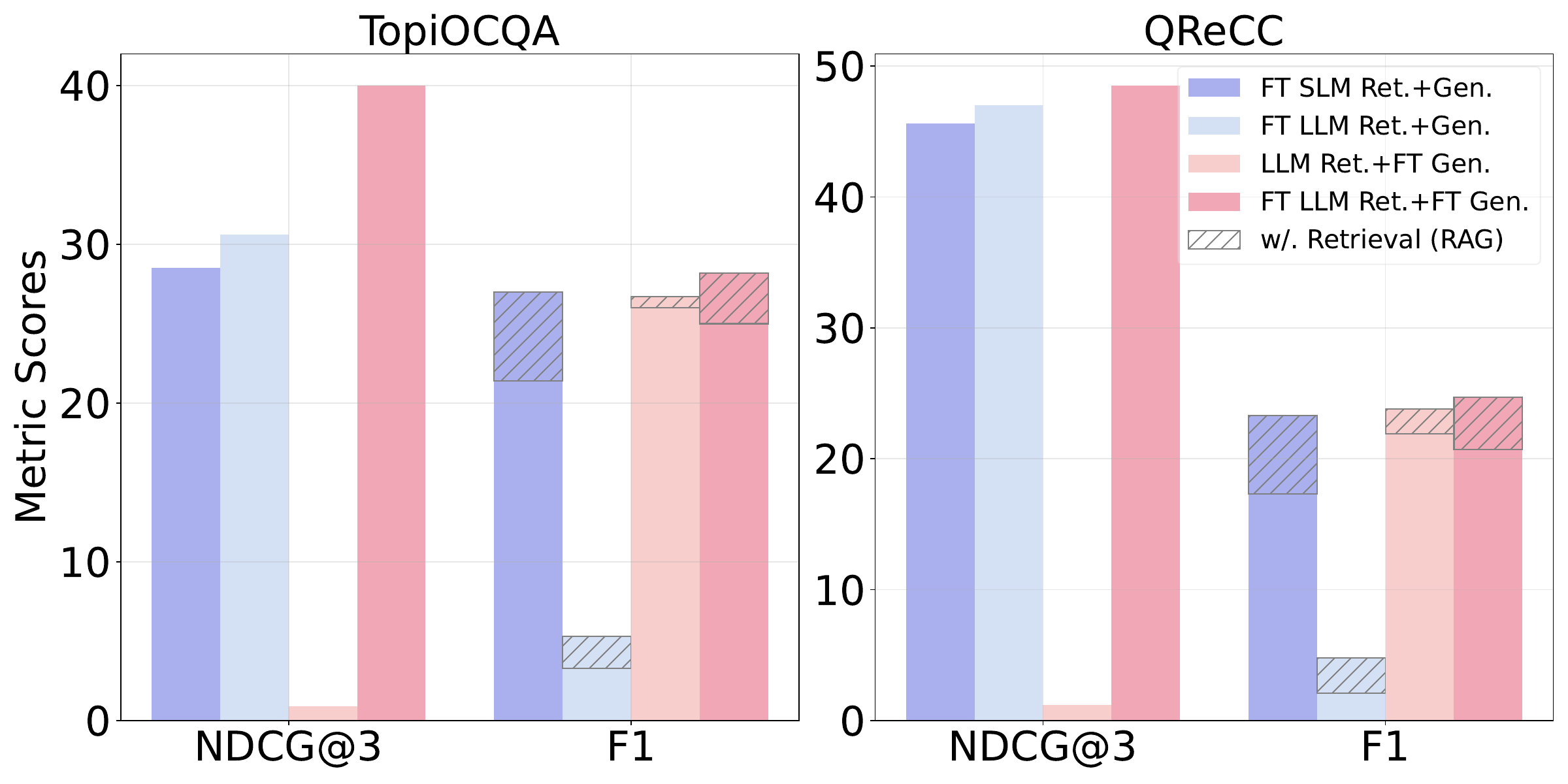}
\caption{The performance of different systems to fine-tune language models for conversational retrieval and response generation with various settings.}
\label{fig: verification}
\vspace{-2ex}
\end{figure}

\subsection{Feasibility of Unifying Conversational Retrieval and Generation}

\begin{table*}[t]
\centering
\scalebox{0.85}{
\begin{tabular}{lcccccccccc}
\toprule
\multirow{2}{*}{Category} & \multirow{2}{*}{System} & \multicolumn{2}{c}{TopiOCQA} & \multicolumn{2}{c}{QReCC} & \multicolumn{2}{c}{OR-QuAC} & \multicolumn{2}{c}{INSCIT} \\
\cmidrule(lr){3-4}\cmidrule(lr){5-6}\cmidrule(lr){7-8}\cmidrule(lr){9-10}
~ & ~ & {NDCG@3} & R@10 & {NDCG@3} & R@10 & {NDCG@3} & R@10 & {NDCG@3} & R@10 \\
\midrule
\multirow{5}{*}{CQR} & \multicolumn{9}{c}{\cellcolor[HTML]{fff8f8}{LLM-based Conversational Query Rewriter (7B) + Ad-hoc Dense Retriever (110M)}} \\
% \midrule
\cmidrule(lr){2-10}
& LLM-Aided & - & - & 41.3 & 65.6 & - & - & - & - \\
& LLM4CS & 26.7 & 43.3 & 42.1 & 66.4 & - & - & - & -\\
& RETPO \small(w./ FT) & 28.9 & 49.6 & 41.1 &  66.7 & - & - & - & -\\
& CHIQ & 32.2 & 53.0 & 44.6 & 70.8 & - & - & - & -\\
\midrule
\multirow{10}{*}{CDR} & \multicolumn{9}{c}{\cellcolor[HTML]{fff8f8}{SLM-based Encoder-only Dense Retriever (110M)}} \\
%\midrule
\cmidrule(lr){2-10}
& ConvDR & 26.4 & 43.5 & 35.7 & 58.2 & - & - & - & -\\
& Conv-ANCE & 28.5 & 52.6 & 45.6 & 71.5 & 35.5 & 55.6 & 24.5 & 38.2 \\
& QRACDR & 36.5 & 57.1 & 49.1 & 74.8 & 40.8 & 60.4 & 30.0 & 43.6\\
%\midrule
\cmidrule(lr){2-10}
& \multicolumn{9}{c}{\cellcolor[HTML]{fff8f8}{LLM-based Decoder-only Dense Retriever (7B)}} \\
%\midrule
\cmidrule(lr){2-10}
& RepLLaMA & 15.0 & 27.2 & 31.8 & 20.4 & - & - & - & -\\
& E5 & 16.9 & 28.7 & 32.9 & 21.1 & - & - & - & -\\
& GRIT & 17.3 & 30.9 & 33.5 & 23.6 & - & - & - & -\\
& Conv-GRIT & 36.0 & 54.2 & 48.3 & 69.7 & - & - & - & -\\
& ChatRetriever & 40.1 & 63.7 & \textbf{52.5} & \textbf{75.8} & 41.9 & 58.9 & 35.1 & 50.8 \\
& UniConv (Ours) & \textbf{42.6}$^\dagger$ & \textbf{67.4}$^\dagger$ & 47.6 & 68.9 & \textbf{43.5}$^\dagger$ & \textbf{63.0}$^\dagger$ & \textbf{36.2}$^\dagger$ & \textbf{54.2}$^\dagger$ \\
\bottomrule
\end{tabular}}
\caption{Performance of different systems for conversational retrieval on four datasets. $\dagger$ denotes significant improvements with t-test at $p<0.05$ over each of the compared CDR systems. \textbf{Bold} indicates the best results.}
\label{table: retrieval}
\vspace{-2ex}
\end{table*}

\noindent We first examine the feasibility of unifying conversational retrieval and generation to verify whether jointly fine-tuning can maintain the model's generation ability while grasping retrieval capacity.
The results for conversational retrieval and response generation on two datasets are shown in Figure~\ref{fig: verification}, which includes four different systems: \textit{i)} fine-tuning SLM for retrieval and using the original LLM as the generator; \textit{ii)} fine-tuning LLM for retrieval only and \textit{iii)} for response generation only; \textit{iv)} fine-tuning LLM for both tasks. 
%Among them, systems i) and iii) are based on two separate models, while systems ii) and iv) are based on one.

By comparing systems ii), iii), and iv), we observe that only fine-tuning a single part on the backbone LLM for retrieval or response generation hurts another ability. However, jointly fine-tuning the model with the objective functions for both tasks can obtain a unified model. %which even mutually improves the performance of the two parts. 
Besides, the LLM-based retriever performs better than the SLM-based one, indicating the potential for conversational search performance with an LLM. 

Then we investigate the RAG setting by incorporating the corresponding retrieved passages for the response generation, except applying the search results from system ii) to system iii), since these two systems cannot handle both tasks.
We can see the improvement from RAG is higher in system iv) with a unified model compared with system iii) with a separated one. These results confirm the feasibility of developing a unified model for conversational search. 
In the following sections, we conduct experiments to investigate our approaches.

% LLM better than SLM, single ft will destroy another ability, unify can mutually improve each other, RAG can improve F1
% Confirm the feasibility then conduct our further design.

\subsection{Results of Conversational Retrieval}
Table~\ref{table: retrieval} shows the conversational retrieval results on four datasets and the comparison with existing systems, where we have the following observations:

(1) Our proposed UniConv outperforms the baseline methods on most of the datasets, including the previous unified model (Conv-GRIT), the state-of-the-art conversational dense retriever (QRACDR and ChatRetriever), and conversational query rewriter (LLM4CS and CHIQ), which demonstrates that the superior dense retrieval ability of our developed system by arousing the LLM capacity with specific fine-tuning. 

(2) The state-of-the-art CDR systems, either SLM-based (QRACDR) or LLM-based (ChatRetriever and UniConv) consistently perform better than the LLM-based CQR systems, which indicates the end-to-end optimization can achieve better performance compared with the rewrite-then-retrieval paradigm~\cite{elgohary2019can}.

(3) The LLM-based retrievers (RepLLaMA, E5) do not always behave much more powerfully than SLM-based ones for conversational retrieval, although they are considered with strong foundational multi-turn context understanding capacity. 
This might be attributed to the possible reason that they are fine-tuned solely on templated instructions, which fail to handle complex and diverse conversational information-seeking scenarios via fully leveraging the generalization capabilities of LLMs. Thus, it is still necessary and important to conduct conversational dense retrieval fine-tuning when employing LLM as a backbone model.

\subsection{Results of Conversational Response Generation}
% zero shot and rag
Table~\ref{table: generation} shows the results of conversational response generation on four datasets with two different settings and the comparison among existing systems. 
In the zero-shot scenario, our UniConv does not perform as well as the current state-of-the-art LLM. This suggests that joint fine-tuning to enhance retrieval capabilities may negatively impact direct response generation performance based on parametric knowledge, due to modifications to the model parameters.
In the RAG setting, where responses are generated based on retrieved passages, we observe that our UniConv outperforms the compared systems with separate retrievers and generators. This indicates that the unified framework may better leverage intrinsic consistency and shared knowledge, mutually enhancing the performance of both retrieval and generation.
% compared with GRIT
\begin{table}[h]
\centering
\scalebox{0.79}{
\begin{tabular}{lcccc}
\toprule
{System} & \small{TopiOCQA} & \small{QReCC} & \small{OR-QuAC} & \small{INSCIT} \\
\midrule
\multicolumn{5}{c}{\cellcolor[HTML]{fff8f8}{w/o retrieval (Zero-shot)}} \\
\midrule
Mistral & 26.6 & 24.3 & 17.4 & 23.1\\
Claude & 27.2 & 25.0 & 17.5 & 27.0\\
ChatGPT & 28.5 & 25.5 & 17.8 & 24.4\\
GRIT & 27.5 & 25.2 & 17.0 & 23.6\\
Conv-GRIT & 26.0 & 23.7 & 14.5 & 23.0\\
\midrule
UniConv & 26.7 & 21.2 & 12.6 & 23.8\\
\midrule
\multicolumn{5}{c}{\cellcolor[HTML]{fff8f8}{w/. retrieval (RAG)}} \\
\midrule
\small Conv-ANCE + Mis. & 27.2 & 25.9 & 17.0 & 24.8\\
\small ChatRetriever + Mis. & 28.3 & \textbf{26.3} & 17.3 & 30.3\\
\small Conv-GRIT & 28.8 & 26.0 & - & - \\
\midrule
UniConv & \textbf{29.6} & 26.2 & \textbf{17.8} & \textbf{33.2} \\
\bottomrule
\end{tabular}}
\caption{Performance of different systems for conversational response generation. For RAG, we use Mistral-7B-chat as the generator to make the results comparable, except for the Conv-GRIT with the same workflow as our UniConv. \textbf{Bold} indicates the best results.}
\label{table: generation}
\vspace{-2ex}
\end{table}

\begin{table}[t]
\centering
\scalebox{0.85}{
\begin{tabular}{lcccc}
\toprule
{Ablation} & \small{TopiOCQA} & \small{QReCC} & \small{OR-QuAC} & \small{INSCIT} \\
\midrule
\small Our UniConv & 42.6 & 46.6 & 43.5 & 36.2 \\
\small \quad w/o CII & 45.5 & 49.7 & 47.6 & 40.0\\
\small \quad w/o DDM & 41.5 & 45.4 & 41.1 & 35.2\\
\bottomrule
\end{tabular}}
\caption{Results of ablation studies for conversational retrieval on four datasets with NDCG@3 score.}
\label{table: ablation_retrieval}
%\vspace{-2ex}
\end{table}

\begin{table}[t]
\centering
\scalebox{0.8}{
\begin{tabular}{lcccc}
\toprule
{System} & \small{TopiOCQA} & \small{QReCC} & \small{OR-QuAC} & \small{INSCIT} \\
\midrule
\multicolumn{5}{c}{\cellcolor[HTML]{fff8f8}{w/o retrieval (Zero-shot)}} \\
\midrule
Our UniConv & 26.7 & 21.2 & 12.6 & 23.8\\
\quad w/o CII & 25.2 & 20.6 & 12.4 & 23.0\\
\quad w/o DDM & 24.8 & 20.8 & 12.3 & 23.7\\
\midrule
\multicolumn{5}{c}{\cellcolor[HTML]{fff8f8}{w/. retrieval (RAG)}} \\
\midrule
Our UniConv & 29.6 & 26.2 & 17.8 & 33.2 \\
\quad w/o CII & 29.4 & 26.0 & 17.3 & 31.4\\
\quad w/o DDM & 29.1 & 24.7 & 16.8 & 25.3\\
\midrule
\multicolumn{5}{c}{\cellcolor[HTML]{fff8f8}{For Reference (Optimal retrieved results)}} \\
\midrule
\quad w/. gold & 41.1 & 26.9 & 23.3 & 34.6\\
\bottomrule
\end{tabular}}
\caption{Results of ablation studies for conversational response generation in two settings with F1 scores.}
\label{table: ablation_generation}
%\vspace{-2ex}
\end{table}

\begin{table*}[t]
\centering
\scalebox{0.8}{
\begin{tabular}{lcccccccc}
\toprule
\multirow{2}{*}{System} & \multicolumn{4}{c}{TopiOCQA} & \multicolumn{4}{c}{FaithDial}\\
\cmidrule(lr){2-5}\cmidrule(lr){6-9}
& F1 (r',r) & Bert (r',r) & F1 (r', E) & Bert (r', E) & F1 (r',r) & Bert (r',r) & F1 (r', E) & Bert (r', E)\\
\midrule
Separated & 23.8 ($\uparrow$ 2.9) & 86.0 ($\downarrow$ 0.3) & 25.5 ($\uparrow$ 2.6) & 87.0 ($\downarrow$ 1.1) & 11.4 ($\uparrow$ 0.7) & 85.0 ($\uparrow$ 10.4) & 10.9 ($\uparrow$ 3.9) & 84.3 ($\uparrow$ 9.5) \\
Unified & 26.7 ($\uparrow$ 2.9) & 86.5 ($\uparrow$ 0.5) & 25.1 ($\uparrow$ 6.9) & 87.4 ($\uparrow$ 0.7) & 11.6 ($\uparrow$ 0.9) & 85.5 ($\uparrow$ 9.8) & 12.1 ($\uparrow$ 4.2) & 87.4 ($\uparrow$ 7.8)\\
\bottomrule
\end{tabular}}
\caption{The performance comparison on two datasets between the system with separated models for retrieval and generation and the unified ones. The evaluation is conducted between the generated response $r^{\prime}$ with the ground-truth response $r$ and the evidence $E$. Arrows denote the change in the results by incorporating RAG.}
\label{table: reliablity}
%\vspace{-2ex}
\end{table*}

\begin{figure*}[t]
\centering
\includegraphics[width=1\linewidth]{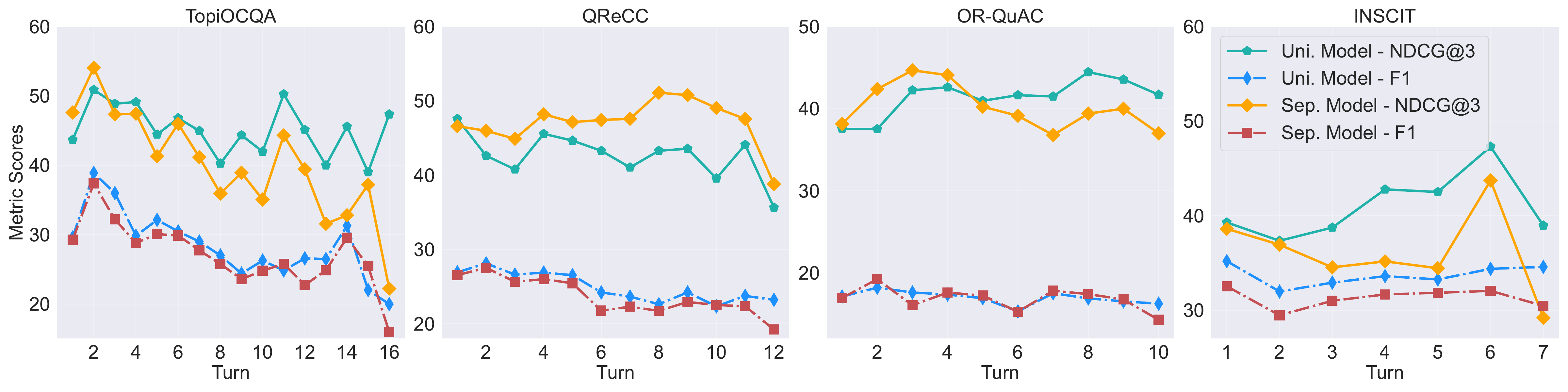}
\caption{The performance of unified (Uni.) and separated (Sep.) models on dense retrieval (NDCG@3) and response generation (F1) with different conversation turns on four different datasets.}
\label{fig: turn_scores}
\vspace{-2ex}
\end{figure*}

\subsection{Ablation Studies}
We conduct ablation studies for conversational retrieval and response generation to study the effects of our proposed two mechanisms, a context identification instruction (CII) mechanism to improve consistency when leveraging the retrieved information for response generation within the same model and a data discrepancy mitigation (DDM) mechanism to induce well-formatted training data with desirable supervision signals. The results are reported in Table~\ref{table: ablation_retrieval} and Table~\ref{table: ablation_generation}, respectively. \\

\noindent \textbf{Conversational Retrieval.} Table~\ref{table: ablation_retrieval} shows an interesting phenomenon that incorporating the CII mechanism would hurt the retrieval performance of our UniConv, while it is helpful for the response generation as shown in Table~\ref{table: ablation_generation}.
This might be because the changed input query form as  $q_n^{\prime} \circ p_n^+$ inevitably influences the contextualized embedding obtained via the learning objective of retrieval $\mathcal{L}_R$, leading to the possible confusion within the model due to the training is conducted simultaneously. A potential solution is to perform fine-tuning for conversational retrieval $\mathcal{L}_R$ and context identification instruction (CII) $\mathcal{L}_{\text{CII}}$ separately into a two-stage process, which can be explored in future studies.
Furthermore, removing the data discrepancy mitigation (DDM) mechanism leads to performance degradation, indicating that utilizing well-structured conversational search data with distinct ground-truths for the retrieval and generation stages during joint fine-tuning can enhance previously sub-optimal results. \\

\noindent \textbf{Conversational Response Generation.} 
Table~\ref{table: ablation_generation} shows that removing any mechanism leads to performance degradation for both zero-shot and RAG settings. 
These observations validate the effectiveness of the added components in enhancing model performance by addressing inconsistencies between retrieval and generation within the unified model.
The improvements vary across datasets, suggesting that the effectiveness of the added mechanisms may depend on the structure and distribution of the data.
Additionally, an obvious gap remains compared to using gold evidence for generation, indicating the potential for further improvement in better integrating retrieved information with the generation process.

\subsection{Impact of Generated Response Reliability}
\label{sec: reliability}
In this section, we investigate whether the unified model can produce a more accurate and reliable response than the system with the separated models. 
We use the variants of UniConv without adding the $\mathcal{L}_{\text{CII}}$ term in Eq.~\ref{eq: loss} as the generator and employ ChatRetriever as the retriever for the RAG setting within the separated system while deploying the full UniConv as the unified system.
We evaluate both systems on the TopiOCQA and FaithDial datasets, measuring similarity using F1 and BERT scores to assess the accuracy of the generated response $r^{\prime}$
compared to the ground-truth response $r$. Faithfulness is evaluated by comparing the generated responses against the evidence or rationale $E$ provided by the datasets. Since FaithDial does not include a retrieval collection, we utilize the same database as TopiOCQA for this purpose.

The results presented in Table~\ref{table: reliablity}, show that the unified system consistently enhances the accuracy of generated responses across both datasets in two settings. For faithfulness, the RAG setting further improves the unified system's performance, whereas a performance drop is observed for the separated system in TopiOCQA.
These observations suggest that developing the system as a unified model can improve reliability to a certain extent.

\begin{table}[t]
\centering
\scalebox{0.87}{
\begin{tabular}{lcccc}
\toprule
{System} & \small{TopiOCQA} & \small{QReCC} & \small{OR-QuAC} & \small{INSCIT} \\
\midrule
\multicolumn{5}{c}{\cellcolor[HTML]{fff8f8}{w/. historical top-3 search results}} \\
\midrule
Separated & 30.3 & 25.3 & 17.2 & 32.0\\
Unified & 31.1 & 26.6 & 18.3 & 33.5 \\
\bottomrule
\end{tabular}}
\caption{The response generation performance comparison by investigating the history-aware ability of different types of systems with F1 scores.}
\label{table: History-Aware}
\vspace{-2ex}
\end{table}

%The results are reported in Table~\ref{table: reliablity}, where the unified system consistently improves the accuracy of generated response in two datasets both for zero-shot and RAG settings. For faithfulness, the RAG further boosts the performance of the unified system, while we observe a drop for the separated system in TopiOCQA.
%These observations confirm the reliability could be improved to some extent by developing the system with a unified model.

\subsection{Impact of Conversational Context}
We examine the impact of conversational context (multi-turn conversations) on retrieval and response generation tasks for systems with unified and separated models. The evaluation is based on per-turn performance, with the implementation for both systems consistent with the setup described in Sec.~\ref{sec: reliability}.
As shown in Figure~\ref{fig: turn_scores}, the unified model consistently outperforms the separated model on both tasks as the conversation progresses, except for the retrieval task on QReCC. This observation highlights the unified model's more robust ability to understand conversations and maintain better consistency between retrieved results and its augmented generation, even in longer conversations.

\subsection{Impact of History-Aware Ability}
We analyze the history-aware ability of the developed model by incorporating the top-3 search results from each historical turn for the current turn's response generation, since the existing studies~\cite{Pan2024Conv-CoA,ye2024boosting} demonstrate that useful information should be contained in history. To ensure a fair comparison, we use the same search results from our UniConv for both systems, varying only the generators as the previous sections.
The results shown in Table~\ref{table: History-Aware} indicate the better performance of the unified model, which suggests the implicit de-noising capacity could be enhanced via the jointly fine-tuning.
This observation also implies that more advantages are still to be discovered within the unified framework.

\section{Conclusion and Future Work}
In this paper, we present UniConv, a unified LLM-based model capable of handling both dense retrieval and response generation in complex conversational scenarios. We propose two mechanisms to seamlessly integrate retrieved information into response generation and address data discrepancy issues during joint fine-tuning. Experimental results on five conversational search datasets demonstrate the superior performance and enhanced reliability of UniConv.
For future studies, developing a unified system for a broader range of complex conversational search scenarios is valuable, including product search, item recommendation, proactive retrieval, etc. 
Besides, it is important to continue improving the consistency between retrieval and generation and conduct specific training based on large-scale synthetic data.
% generate better data

\section*{Limitations}
Despite our comprehensive studies, some potential limitations can be addressed in the future:

\noindent \textbf{Efficiency.} 
The used backbone model with 7B size LLM is larger than the previous SLM-based CDR systems, which raises efficiency concerns. 
Nevertheless, on the one hand, the LLM-based retriever with superior search performance reduces the requirement for extensive passage re-ranking.
In real-world applications, this could help reduce the initial higher costs by ultimately decreasing the overall time required for ranking.
On the other hand, the multi-task ability of UniConv makes the cost worthwhile compared with the retrieval-only systems in existing studies. This is also a promising research direction that integrates more embedding-based tasks into the instruction-based generation framework in conversation. 
Besides, exploring the possibility of distilling UniConv into a more efficient, smaller model is desirable. \\

% better hyper-parameters and data ratio/type, backbone
\noindent \textbf{Broader Experimental Configuration.}
We only leverage the fixed hyper-parameters for model setup and ratio to mix training data. Though we obtain strong performance, the exploration within broader experimental configurations could lead to better performance. Besides, adapting our methods to other types or sizes of backbone models and using full-model fine-tuning rather than LoRA might bring additional observations and results. \\

\noindent \textbf{Robust Evaluation for Generation.}
%broad evaluation aspects (metrics) and backbone
How to evaluate the generated content is still an open question for the research community. Our evaluation is conducted on a single metric following the previous studies~\cite{mao2024chatretriever,liu2024chatqa} for a fair comparison, which might not reflect the quality of different aspects of the generated response. Leveraging more comprehensive evaluation metrics or incorporating another LLM as an evaluator might help us to observe more insights about improving the consistency between retrieval and generation.

%\section*{Acknowledgments}
%...

\bibliography{custom}

\appendix
\section*{Appendix}
\section{Experimental Setup}
\label{sec: appendix_experiment}

\subsection{Datasets Details}
\label{sec: appendix_dataset}
\begin{table}[h]
\centering
\scalebox{0.8}{
\setlength{\tabcolsep}{4pt}{
\begin{tabular}{lrrrrr}
\toprule
 & \small TopiOCQA & \small QReCC & \small OR-QuAC & \small INSCIT & \small FaithDial\\ 
\midrule
\small\#Conv. & 205 & 2,775 & 771 & 469 & 791\\
\small\#Turns(Qry) & 2,514 & 16,451 & 5,571 & 2,767 & 3,539\\
\small\#Collection & 25M & 54M & 11M & 49M & -\\
\small\#Avg. Qry & 12.9 & 5.3 & 7.2 & 5.9 & 4.5\\
\small\#Min Qry & 5 & 2 & 4 & 2 & 4 \\
\small\#Max Qry & 25 & 12 & 12 & 7 & 5\\
\small\#Avg. Psg & 9.0 & 1.6 & 1.0 & 1.6 & -\\
\bottomrule
\end{tabular}}}
\caption{Statistics of five used datasets.}
%\vspace{-2ex}
\label{table: datasets}
\end{table}
The statistics of each dataset are presented in Table~\ref{table: datasets}. The first four datasets are used for the retrieval and response generation evaluation while the FaithDial does not provide the collection for retrieval so it is used for reliability evaluation only.

\subsection{Baseline Details}
\label{sec: appendix_baseline}
We provide a more detailed introduction to the following baselines used for comparison:

\textbf{LLM-Aided}~\cite{ye2023enhancing}: An informative conversational query rewriting by directly prompting ChatGPT-3.5 as both query rewriters and rewrite editors twice to incorporate all the desirable properties for producing the final rewritten queries.

\textbf{LLM4CS}~\cite{mao2023large}: A state-of-the-art LLM-based prompting method for conversational query rewriting. We implement it with full aggregation by calling LLMs five times for query and response generation but without the chain-of-thought (CoT) content because of the efficient annotation consideration in practical scenarios.

\textbf{RETPO}~\cite{yoon2024ask}: A retriever preference adapted query rewriting method that fine-tunes LLaMA-2-7B-Chat as a query rewrite model with an external query rewrite dataset generated by GPT-4.

\textbf{CHIQ}~\cite{mo2024chiq}: A state-of-the-art method leverages the basic NLP capabilities of LLMs to enhance the quality of contextual history for improving the query rewriting performance.

\textbf{ConvDR}~\cite{yu2021few}: A conversational dense retrieval method that uses knowledge distillation to learn the session embeddings with relevance judgments from the human-rewritten queries based on the ANCE model.

\textbf{Conv-ANCE}~\cite{mao2023learning}: A conversational dense retrieval method that leverages ANCE fine-tuned on conversational search data only using the retrieval loss term in Eq.~\ref{eq: loss}.

\textbf{QRACDR}~\cite{mo2024aligning}: A state-of-the-art SLM-based query representation alignment conversational dense retrieval method by incorporating relevance judgments and rewritten query annotation as supervision signals for retriever fine-tuning.

\textbf{RepLLaMA}~\cite{ma2024fine}: A large ad-hoc dense retriever fine-tuned on top of the LLaMA-7B-Chat model on the MSMARCO dataset.

\textbf{E5}~\cite{wang2024improving}: A large ad-hoc retriever fine-tuned on top of Mistral-7B model on the combination of synthetic dataset generated by ChatGPT-3.5 and MSMARCO.

\textbf{CharRetriever}~\cite{mao2024chatretriever}: A state-of-the-art LLM-based conversational dense retriever with better robustness via a novel contrastive session-masked instruction tuning approach.

\textbf{GRIT}~\cite{muennighoff2024generative}: A first proposed unified model to handle both retrieval and generation tasks by incorporating vanilla instruction tuning and using different training data for its contrastive learning and instruction tuning.

\textbf{Conv-GRIT}~\cite{muennighoff2024generative}: A variant of GRIT fine-tuned on the conversational data with the same setting as our UniConv model for fair comparisons.

\end{document}